\documentclass[journal]{IEEEtran}

\ifCLASSINFOpdf
   \usepackage[pdftex]{graphicx}
\else
\fi
\usepackage{amsmath}
\usepackage[byname]{smartref}
\usepackage[capitalise]{cleveref}
\usepackage[linesnumbered,ruled]{algorithm2e}
\usepackage{tabularx}
\usepackage{booktabs}
\usepackage{amssymb}

\begin{document}
%
\title{
Appliance Operation Modes Identification Using Cycles Clustering}
%
%
%

\author{Abdelkareem Jaradat, Hanan Lutfiyya, Anwar Haque
	\\\{ajarada3,  hlutfiyy, ahaque32\}@uwo.ca
	\\The University Of Western Ontario -
	Department Of Computer Science
	\\London, Ontario, Canada
    }

%
%

\markboth{ieee transaction on smart grid}%
{Shell \MakeLowercase{\textit{et al.}}: Bare Demo of IEEEtran.cls for IEEE Journals}
%



\maketitle

\begin{abstract}
The increasing cost, energy demand, and environmental issues has led many researchers to find approaches for energy monitoring, and hence energy conservation. The emerging technologies of Internet of Things (IoT) and Machine Learning (ML) deliver techniques that have the potential to efficiently conserve energy and improve the utilization of energy consumption.  Smart Home Energy Management Systems (SHEMSs) have the potential to contribute in energy conservation through the application of Demand Response (DR) in the residential sector. In this paper, we propose appliances Operation Modes Identification using Cycles Clustering (OMICC) which is SHEMS fundamental approach that utilizes the sensed residential disaggregated power consumption in supporting DR by providing consumers the opportunity to select lighter appliance operation modes. The cycles of the Single Usage Profile (SUP)  of an appliance are  extracted and reformed into  features in terms of clusters of cycles. These features are then used to identify the operation mode used in every occurrence using K-Nearest Neighbors (KNN). Operation modes identification is considered a basis for many potential smart DR applications within SHEMS towards the consumers or the suppliers.
\end{abstract}


%
\IEEEpeerreviewmaketitle

\section{Introduction}
\IEEEPARstart{t}{he} conservation of non-renewal resources is of interest not only because these resources are not renewable in the face of accelerated demand by the industrial and residential sectors, but also because of the impact on the climate \cite{jorde2020event}. Extensive research is currently  being carried out on finding solutions to efficiently use energy sources. Research suggests that involving the consumer in energy conservation through insights on the details of energy consumption is effective and creates a noticeable difference in the reduction of energy waste \cite{armel2013disaggregation}. Such involvement preferably requires providing consumers with individual load-level consumption information (diaggregated) rather than only providing the mains consumption information (aggregated) that represents the total consumption of all loads \cite{Pouresmaeil}. Exposing individual appliance-level energy information is capable of reducing energy consumption by increasing the consumer's awareness of the consumption of individual appliances. \cite{kelly2016does}. Disaggregated consumption information can be obtained with the assistance of smart meters and sensors which can measure per appliance electricity consumption reading, or by decomposing the aggregated mains signal into individual appliance signals using event detection algorithms through Non-Intrusive Load Monitoring (NILM) methods \cite{wang2018review}. 

A mechanism that is popular in many studies, Demand Response (DR), provides an opportunity for consumers to contribute in the reduction in electricity consumption by reducing their consumption or by shifting the electricity usage to off-peak periods in response to time-based tariffs, by either monetary or non-monetary incentives \cite{li2017distributed}. DR may be used to increase demand during periods of high supply and low demand, or DR could be used together with proper feedback to the consumer to reduce the power consumption per appliance. DR  is also considered a more cost-effective option than investing in more power supplies such as building extra power stations \cite{golz2017does}. This is not a new concept since it has been extensively investigated in recent years as an approach to reducing power consumption \cite{li2017distributed}. However, in most countries DR still plays a very limited role in energy conservation \cite{iea}. This is due to the fact that DR requires buildings to be equipped with sensors and automation systems which has still not been widely adopted  \cite{pallonetto2019demand}.

Smart Home Energy Management Systems (SHEMSs) have the potential to connect pieces together and provide a solid solution \cite{alhammadi2019survey} by enabling a set of sustainable smart applications for energy savings [4]. SHEMSs are capable of providing visual feedback to the consumer in the form of energy usage data, automation and control initiated by the utility party, load forecasting, and optimized load scheduling horizon \cite{liu2016review}.  
With smart meters and sensors that can measure power consumption per appliance, it has become much easier with the assistance of signals processing mechanisms to monitor individual device power consumption and support SHEMSs applications  \cite{Yang2016}. In this paper, we propose Operation Modes Identification using Cycles Clustering (OMICC) that is considered a fundamental machine learning approach for SHEMSs to be built on top of OMICC. The main objective of OMICC is to process sensed disaggregated power consumption readings, and identify for each appliance what is the operation mode used for each detected activation time \cite{jaradat2020demand}. During the use of an appliance, there are several operation modes the user can select from. An operation mode is a specific setting set by a manufacturer to meet certain needs for the customer, such as light and heavy modes in a washing machine. Activating an appliance in a certain operation mode consumes energy differently than others. Thus, a DR opportunity arises such that OMICC identifies the operation modes and reports them with activation times to a SHEMS so that the consumer can take a proper action towards energy use reduction. These actions are typically in the form of load shifting and reduction \cite{pallonetto2019demand}. Load shifting is performed by changing the time of use to off-peak hours when the demand is low and the price is cheaper. On the other hand, load reduction is obtained when consumers change their habits in terms of the pattern and frequency of selecting certain operation modes over others. Consequently, selecting lighter operation modes instead of the heavier ones upon appliance usage significantly reduces the power consumption associated with these appliances hence supporting DR. This way, OMICC is considered the analytical layer that a SHEMS relies on to apply DR for the residential sector.

The rest of the paper is organized as the following: 
In section II we discuss previous work in the literature. 
Section III discusses the data analysis of appliances power consumption. 
In section IV we present the proposed approach, OMICC. 
Section V discusses the results and 
section IX concludes the paper and suggests future work.

\section{Related Work}
\label{related}
Demand-response methods are potential solutions for improving the efficiency of future smart power grids. The literature consists of many DR approaches that aim to reduce consumer power bills and decrease the load on the grid.
Haben el.al.\cite{haben2015analysis} proposed a clustering method using Finite Mixture Models (FMMs) to identify households that are more suitable for demand reduction, and discovering clusters that models the types of peak demands and major any seasonality and variation in the data.
Zheng et.al \cite{zheng2020incentive} proposed an incentive-based model for multiple energy carriers that takes into account the behavioral coupling effect of the consumers and the impact of energy storage unit. The results confirm benefits of the proposed model in reducing cost of multi-energy aggregator, and reduction in the dissatisfaction of consumers. A comprehensive optimization-based Automated Demand Response  controller \cite{althaher2015automated} is developed to optimize the operation of several types of household appliances aiming for reduction in consumer’s power bill within a predefined range with dissatisfaction minimization.
A real-time price-based DR algorithm \cite{yu2015real} is presented to gain optimal load control of devices in certain facility by creating a virtual electricity-trading process that utilizes the Stackelberg game.
An advanced reward system  \cite{hu2016framework} presented the concept of a comfort indicator in a framework where demand reduction requests for household appliances is communicated efficiently maintaining households’comfort levels.
A demand response optimization framework \cite{li2017distributed} from the utility perspective was developed  to minimize the  user costs and utility cost.

The literature describes many techniques used to analyze the electricity consumption data so that end user applications can be built on top of these approaches \cite{Wang8322199}. These techniques are primarily concerned with event detection \cite{jorde2020event} and event classification \cite{jaradat2020demand} in time series data. Typically the event is the activation and deactivation of appliances during its operation with the power distribution over time. 
Different approaches are concerned with extracting load signatures from aggregated power consumption data and classifying these signatures into the individual loads such as appliances and lights, best known as Non Intrusive Load Monitoring (NILM). 
Liao et.al. \cite{liao2014power} proposed an approach for appliance load identification using Dynamic Time Warping (DTW).
Liu et.al. \cite{liu2017dynamic} used a Nearest Neighbor Transient Identification method to identify the appliance creating the Transient Power Waveform (TPW) sample time-series. The DTW-based integrated distance is then utilized to calculate the similarity of TPW signatures and a template time series for an appliance. 
Wang et.al. \cite{wang2018iterative} describes a NILM approach which uses Iterative Disaggregation based on Appliance Consumption Pattern (ILDACP). This approach combines Fuzzy C-means clustering algorithm to detect appliance operating status, and DTW search that identifies individual appliances energy consumption based on the appliance typical power consumption pattern (a template pattern). 

Machine learning algorithms are emerging in the context of load signature detection with both supervised and unsupervised algorithms. Barsim et.al. \cite{barsim2014approach} proposed an approach that uses the typical event-based NILM system with only unsupervised algorithms to eliminate the need for training stage. The event detector is a grid-based clustering algorithm to segment the power signals into transition-state and steady-state segments. They extracted a set of features from the detected events and used them in a Mean-Shift Clustering algorithm.  Fernandez et. al. \cite{fernandez2016online} performed online learning approach for the identification of home appliances based on the Confidence Weighted algorithm (CW) and six other algorithms. Kang et.al. \cite{kang2016classification} used Probabilistic K-Nearest Neighbor (PKNN) to infer the device states from home appliance electrical power usage signals and also from sensor data that includes temperature and humidity. Prudenzi et.al. \cite{prudenzi2002neuron} proposed a procedure that provides three different sequential back propagation Artificial Neural Networks (ANNs) to process the load shape and identify load signatures. The  classification is then performed by an unsupervised network implementing the Self-Organizing Map (SOM) of Kohonen \cite{prudenzi2002neuron}.

\begin{table}[t]
	\caption{Classification for the electrical loads based on operation behavior and the selected appliances to analyze in this paper.}
	\centering
	\begin{tabularx}{.475\textwidth}{XXX}
		\bottomrule\hline
		\textbf{Load Type	}		& \textbf{Examples} & \textbf{Selection} \\ 
		
		\hline
		Timer Based Loads	 & 
		Air conditioning, water heater, refrigerator & - \\

		Preprogrammed Loads	 & Dishwashers,  Ovens, Broilers, clothes washers, and clothes dryers & Dishwashers, clothes washers, and clothes dryers  \\

		Variable Loads & electric cook-top, lighting, television, hair dryer, laptops, and almost the rest of house appliances & - \\
		\bottomrule\hline
	\end{tabularx}
	\label{table:LoadsClassification}
\end{table}

Most of the work currently in the literature uses (NILM) \cite{costanzo1986energy} to identify individual appliances and loads from an aggregated load signal. The literature is limited in  approaches that focus on analyzing disaggregated power data to identify the activation of household appliances and then classify each use of the appliance into its operation modes. With OMICC, appliance operation modes are identified, and thus, a DR mechanism is possible for consumers such that by lowering the frequency of using heavier operation modes the overall consumption per appliance is reduced.

\section{Electrical Load Data Analysis}
We use Power Consumption DataSets (PCDSs) \cite{Makonin2017} to understand the power consumption characteristics of household electrical loads over time when used and discover potential DR opportunities\cite{Pipattanasomporn2014}. A household electrical load is an electrical device, component or sub-component of a circuit that is used by a house residents, where this load consumes electric active power \cite{wiki:loads}. The main objective of this analysis is to understand the states that an electrical load goes through when these loads are used. This assists in forming statistical models representing each load over time. These models have the potential to replicate this data into a form of synthetic datasets which assist in validating proposed algorithms plus reduce the time and effort in collecting real PCDSs.

\subsection{Electrical Load Classification }
\label{LoadsClassification}
Based on the characteristics of electrical loads, a classification is presented in Table \ref{table:LoadsClassification} that categorizes electrical loads into three categories. \textbf{Timer Based Loads} consist of a set of cycles that is replicated over time while the load is active. These loads are regulated using a timing regulation controller, such as a thermostat that responds to temperature  fluctuations. For example, in a refrigerator, the compressor cools the refrigerator interior and is activated when the interior temperature passes a preset upper threshold, however, it switches off when the temperature falls below a preset lower threshold.
\begin{figure}[t]
	\centering
	\fbox{\includegraphics[trim={ 1.0cm 1.0cm 1.0cm 1.0cm },clip,width=.475\textwidth]{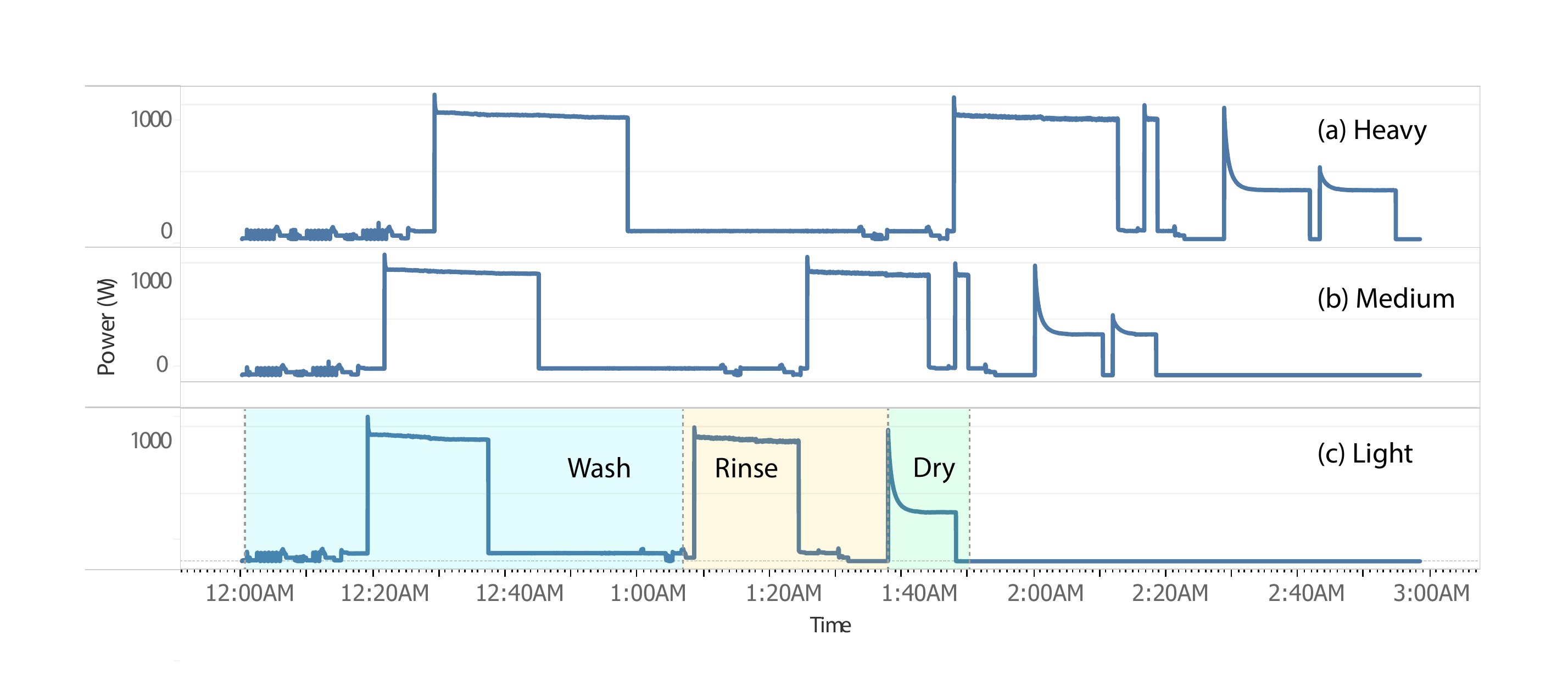}}
	\caption{SUPs for the dishwasher showing three operation modes (a) Heavy, (b) Medium, (c) Light.}
	\label{fig:Dishwasher_All_modes}
\end{figure}
In the \textbf{Variable Loads} category, loads are triggered on and off over time based on human behavior where the timing distribution of the interventions is correlated to a wide range of parameters related to households occupancy and life style such as wake up times, bed times, work times, number of house occupants and their ages, etc. Therefore, without having sufficient information about these parameters, it is difficult to precisely predict the start time nor the working duration for loads belong to this category. Consequently, the electrical load behavior of this kind of loads relies on their usage statistics. 
\textbf{Preprogrammed Loads} are activated manually by human intervention and usually consume relatively large amounts of power in a relatively short time. Once a specific load is activated, it goes through a set preprogrammed states where each state has a specific duration until the load automatically shuts off. A dishwasher is an example of this category. It cycles through a set of states such as water filling, washing, and rinse state. Each state is configured with a duration, and a power level. In this paper, the focus is spotted on analyzing the power consumption characteristics of appliances belong to this category.

\subsection{Single Use Profile and Operation Modes}
A Single Use Profile \textbf{(SUP)} is used to formally model power consumption of a preprogrammed appliance between the time it is turned on and the time it is turned off. SUP represents the sequence of power consumption values consumed by an appliance from the moment of turning it on to the moment of that it is turned off. Hence, SUP with the sampling frequency $f_s=1Hz$  is defined by the sequence $\{ p_i \}_{i=t_{on}}^{t_{off}}$ where $p_i$ is the instantaneous power reading at time $i$. The time stamps $t_{on}, t_{off}$ are the turn on, turn off times respectively of the appliance and $i \in [t_{on}, t_{off}]$ represents the index the $i^{th}$ sample. 
Typically, home appliances may run a SUP in one of several operation modes. An operation mode is characterized by its running time and different cycles that the appliance passes through. For example, a dishwasher may have two operation modes, a lighter one for barely used dishes, and a heavy one for greasy dishes. Activating certain appliance with a SUP in a certain operation mode consumes electricity differently than other operation modes.

\subsection{Analysis Of Individual Appliances}
 In this paper, the focus is on analyzing the behavior of the preprogrammed appliances. The appliances data that we analyzed are the clothes washer, clothes dryer, and dishwasher.

\subsubsection{The Dishwasher}
\begin{figure}[t]
	\centering
	\fbox{\includegraphics[trim={1.7cm 1.7cm 1.7cm 1.7cm },clip,width=.475\textwidth]{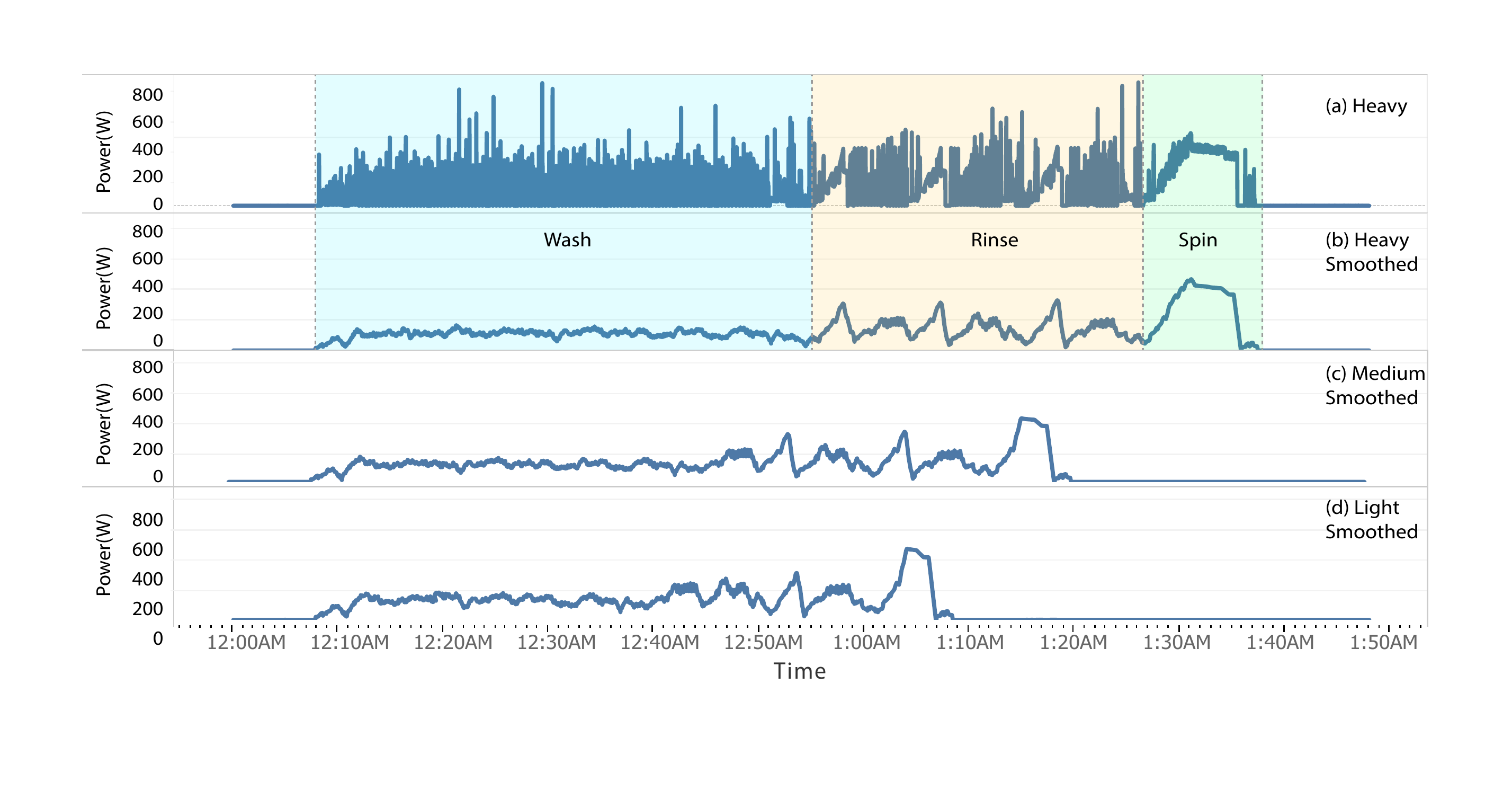}}
	\caption{\textbf{(a)} SUPs for clothes washer using heavy operation mode. \textbf{(b)} Smoothed SUP plot in heavy operation mode. \textbf{(c)} Smoothed SUP plot in medium operation mode. \textbf{(d)} Smoothed SUP plot in light operation mode}
	\label{fig:washer_heavy}
\end{figure}

The dishwasher has three main operation modes: Heavy, Medium, and Light. Each SUP of the dishwasher has three states: wash, rinse and dry, regardless of the operation mode. Three SUPs in three operation modes for the dishwasher are depicted in \cref{fig:Dishwasher_All_modes}. 
The wash state involves filling the dishwasher with water and spraying the water through jets to get the first round of spray with regular water temperature.  The water then is heated to the desired temperature setting.  The sprinkler then starts washing by rotating and spraying hot water on the dishes.  The rinse state that follows repeats the steps in the wash state. Finally, the dry state starts by increasing the interior air temperature of the machine for a short time since the insulator installed in the dishwasher keeps the interior on high temperature until the dry states completes.

\subsubsection{The Clothes Washer}
Clothes washer SUP has three states during operation: wash, rinse, and spin.
\cref{fig:washer_heavy} graphically depicts SUPs of the clothes washer for three operation modes: Heavy, Medium, and Light. \cref{fig:washer_heavy} (a) is the real consumption data plotted over time in heavy operation mode. In \cref{fig:washer_heavy} (b, c, d) the plots corresponds to a smoothed version of the SUP plot using moving median to make it easier to understand.
The washer starts its wash state by filling in the water to the main cavity, during which the washer consumes relatively low power. As the water fills in, the washer starts rotating. Initially, the power consumption is low but gradually increases due to the rise of the motor electrical load as the water level climbs up. Frequent variations in motor speeds are observed during the wash state. The rinse state follows for a shorter time. Finally, the spin state occurs when the machine stops pumping water to the cavity and starts drying clothes by spinning at a very high speed, therefore, the machine consumes the highest power.

\begin{figure}[t]
	\centering
	\fbox{\includegraphics[trim={1.7cm 1.7cm 1.7cm 1.7cm },clip,width=3.4in]{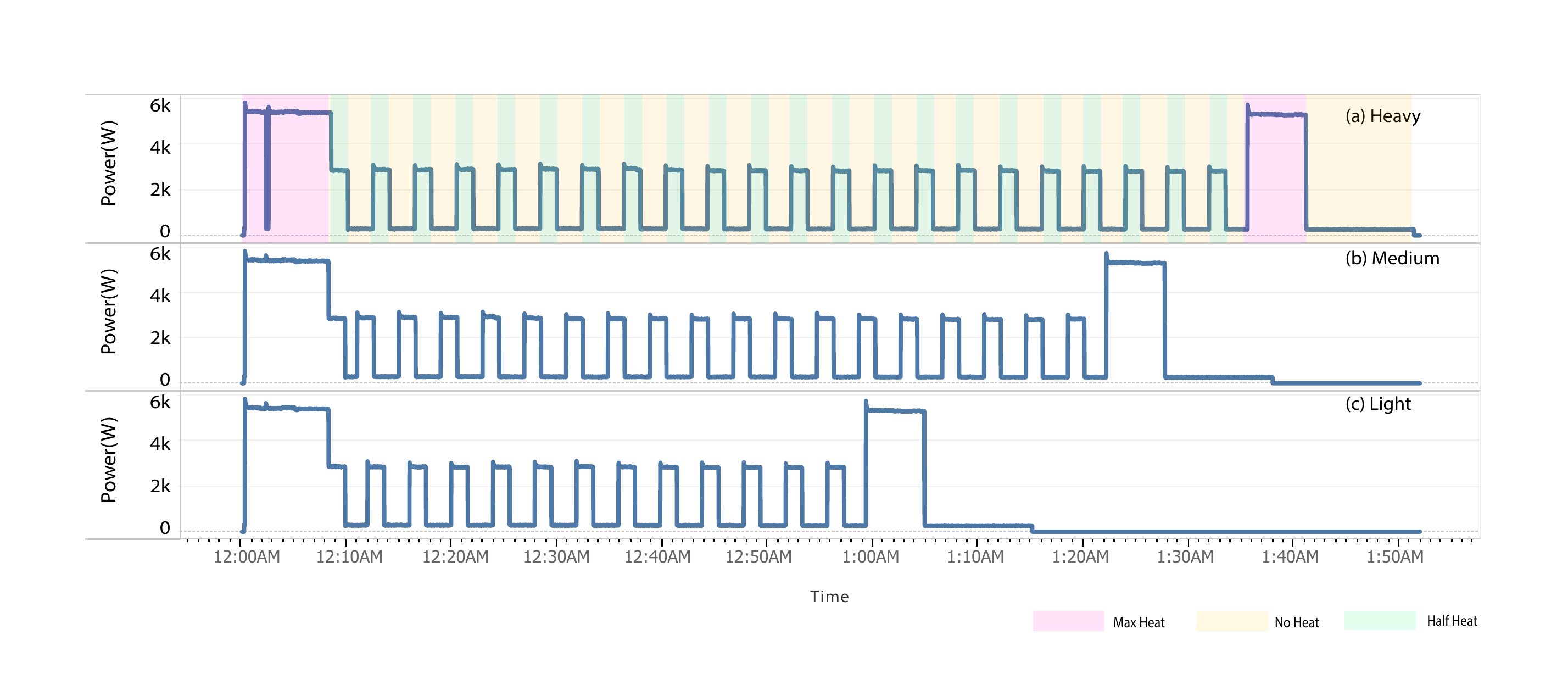}}
	\caption{SUPs for clothes dryer with three operation modes \textbf{(a)} Heavy \textbf{(b)} Medium \textbf{(c)} Light.}
	\label{fig:Dryer_all_modes2}
\end{figure}

\subsubsection{The Dryer}
A SUP of the dryer contains these cycles: maximum temperature, average temperature, and minimum temperature. During the maximum temperature cycle, the dryer drum is heated to its maximum value in the dryer setting, and the duration of this state is relatively longer than other cycles. The average temperature cycle sets the core temperature to about half of the maximum value with relatively short duration. Finally, with the minimum temperature cycle, the heating element in the dryer is turned off.  \cref{fig:Dryer_all_modes2} shows a three SUPs for a dryer in three different operation modes: Heavy, Medium, and Light.

\section{SUP CLASSIFICATION WITH OMICC}
\begin{figure}[t]
	\centering
	\includegraphics[width=3.5in]{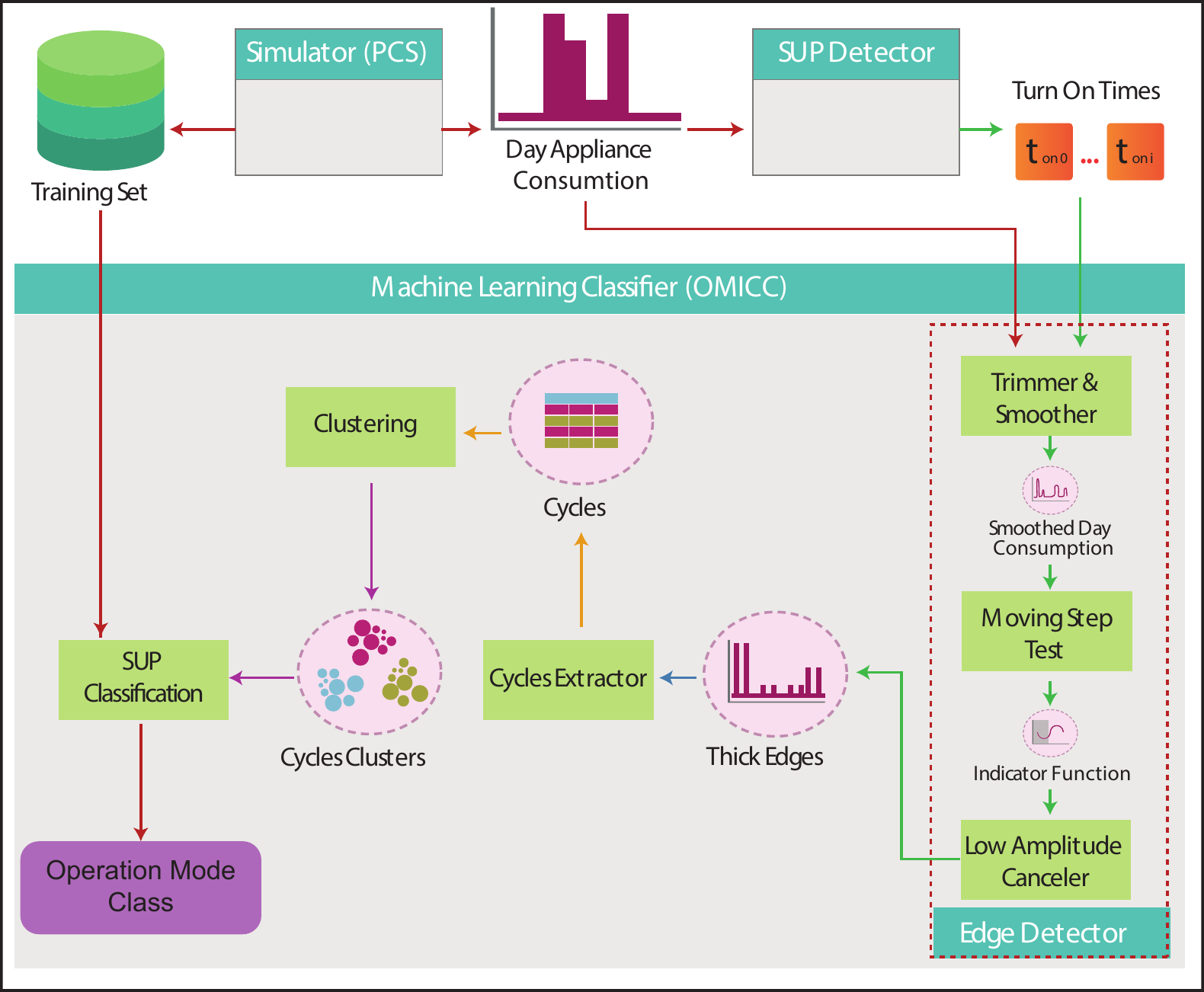}
	\caption{The basic architecture for the classification using OMICC. }
	\label{fig:mlModel}
\end{figure}

We propose  Operation  Modes  Identification using Cycles Clustering (OMICC) approach to classify the operation modes of SUPs. The approach is focused on extracting features of SUPs that serves to classify into operations modes classes. \cref{fig:mlModel} depicts the architecture of the model used in this approach.

\subsection{SUP Features Extraction}
 This section describes the technique for extracting features from a detected SUP assuming a SUP is already detected and the $t_{on}$ is given \cite{jaradat2020demand}. The features are represented as the set of cycles that form a SUP. Each cycle is characterized by two abrupt changes in the power value. These features are used by the classification algorithm to classify the detected SUP into operation modes.

\subsubsection{Edge Detector}	
The classification of the detected SUP into operation modes requires determining the cycles that form the SUP.
The day consumption is represented by $D(t)$. Given a turn on time for a SUP, $ t_{on} $, we define $D'(t)$ as follows:
\begin{equation} \label{eq:ml01}
D'(t) = D(t): \quad t \in [t_{on}, midnight]
\end{equation}
where $  D'(t) $ represents the consumption after the turn on time to the rest of the day. A median smoother function is then applied to $ D'(t) $ to cancel the low amplitude noise component.
Edge detection is based on the  Moving Step-Test (MST) \cite{fried2007robust} where MST is characterized by the following equation:
\begin{equation} \label{eq:ml02}
I(t) = |m_{t+\ell} - m_{t-\ell}|
\end{equation} 
where $I(t)$ is defined as the \textit{Indicator Function}. $I(t)$ is used to detect edges of the cycles by finding a period of time that surrounds each edge where it is highly likely that an edge exists within this period. We define $m_{t+\ell} $  as the median value of the power consumption of the leading sequence that starts at $ D'(t) $ and ends at $ D'(t+\ell) $ and $m_{t-\ell} $ is the median value of the power consumption values for the lagging sequence that starts at $ D'(t-\ell) $ and ends at $ D'(t) $. The median function is selected since it gives a more accurate edge location by eliminating the presence of points that are located at the other end of the transition around the edge.

A cycle in $ D'(t) $ is defined by two abrupt changes in the value of $ D'(t) $. An abrupt change in $ D'(t) $ is a single point of time $t_m \in [t_s,t_e]$ which is referred to as \textit{Exact Edge}. Exact edges are determined based on the values of $I(t)$ such that 
if the value of $I(t)$ is higher than a certain threshold, $\tau$, this indicates the start or the end of a cycle. Each period of time $[t_{s},t_{e}]$ such that $I(t) > \tau$  forms a \textit{Thick Edge} i.e, a thick edge is period of time defined by a starting time  $t_{s}$ and ending time $t_{e}$ where $I(t)$  is greater than $\tau$. A high amplitude of $I(t)$ in a thick edge indicates that the values of the two medians $m_{t-\ell} $ and $m_{t+\ell} $ are far from each other. Which in turns means that an abrupt change (Exact Edge) in $ D'(t) $ exists.

\begin{figure}[t]
	\centering
	\includegraphics[width=3.5in]{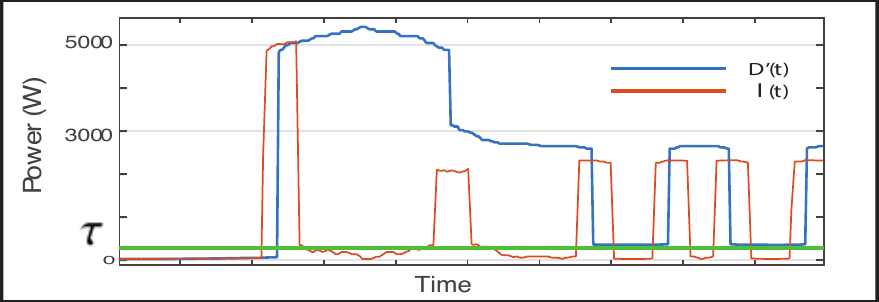}
	\caption{The response of the indicator function $I(t)$ for the abrupt changes in $D'(t)$.}
	\label{fig:mlStepTestZoom}
\end{figure}

The indicator function, $I(t)$, facilitates finding exact edges in $ D'(t) $, rather than looking directly in $ D(t) $.
These sudden changes in $ D'(t) $ occurs at different power levels i.e., an exact edge could occur when the value of $ D'(t) $ is around 5W and jumps to 1000W, or it could change from 2000W to 6000W. 
In addition, exact edges occur from lower values to higher vales (a rising edge)  or vise versa (a falling edge). i.e.,  an exact edge could occur when the power value of $ D'(t) $ is at 5W and rises to 2500W or when the power value falls from 2500W to 5W.
On the other hand, $I(t)$ shows all the exact edges in $ D'(t) $ in  a single reference, such that in the absence of any exact edge in $ D'(t) $, $I(t)$ shows a value closer to zero. However, whenever an exact edge occurs in $ D'(t) $, $I(t)$ shows a spike with high value in a short period of time.

The threshold $\tau$ is calculated based on the standard deviation of $I(t)$ and Low Amplitude Canceling Multiplier, $\zeta$. The value of $\tau$ is defined as follows:

\begin{equation} \label{eq:ml03}
\tau = \zeta . \sigma(I(t)): \quad \zeta \in Z^+
\end{equation} 
where the multiplier, $\zeta$, adjusts the threshold $\tau$ in order to cancel whatever values of $I(t)$ that is less than the threshold value. This is graphically depicted in \cref{fig:mlStepTestZoom}.

By thresholding $I(t)$ with $\tau$, a trimmed version, $I'(t)$, is obtained as presented in \cref{fig:mlCycles}. We define $I'(t)$  as the following:

\begin{equation} \label{eq:ml04}
I'(t) =I(t) - \tau
\end{equation} 
where $I'(t)$  is used to determine the  thick edges set, $E$, within a SUP such that:
\begin{equation} \label{eq:ml06}
\begin{array}{l}
E = \{e_0, e_1, .. , e_{n-1}\} \quad  \\
e_i = [t_{s_i},t_{e_i}]: \quad 0 \leq i < n\\
\end{array}
\end{equation} 
where $n$ is the size of $E$, $t_{s_i}$ and $t_{e_i}$ represent the start time stamp and end time stamp of the $i^{th}$ thick edge respectively. \cref{fig:mlCycles} illustrates the extraction of the thick edges. It shows the plot of $ I'(t) $ and a list of pairs of time stamps. The $i^{th}$ pair defines the start time, $t_{s_i}$, and end time, $t_{e_i}$, of the $i^{th}$ thick edge. 

\subsubsection{Cycles Extractor}
Once thick edges set, $ E$, is obtained, the cycles of the SUP is extracted using the following steps:

\paragraph{Edge Thinning}
This refers to selecting the exact edge $t_{m_i}$ from a thick edge $e_i=[t_{s_i},t_{e_i}]$. We pick $t_{m_i} $ such that it falls exactly in the center of $e_i$, so $t_{m_i} $ is defined as:
\begin{equation} \label{eq:ml05}
t_{m_i}  =  \frac{1}{2}(t_{s_i}+t_{e_i})
\end{equation} 
such that this equation is applied across all thick edges $ e_i \in E$. All thick edges are thinned and grouped in the Exact Edges Set $X = \{t_{m_0}, t_{m_1}, t_{m_2},...t_{m_n},\}$ where $t_{m_i}$ represents the $i^{th}$ exact edge, and $n$ is the number of thick edges, consequently, equals the number of exact edges.

\begin{figure}[t]
	\centering
	\includegraphics[width=3.5in]{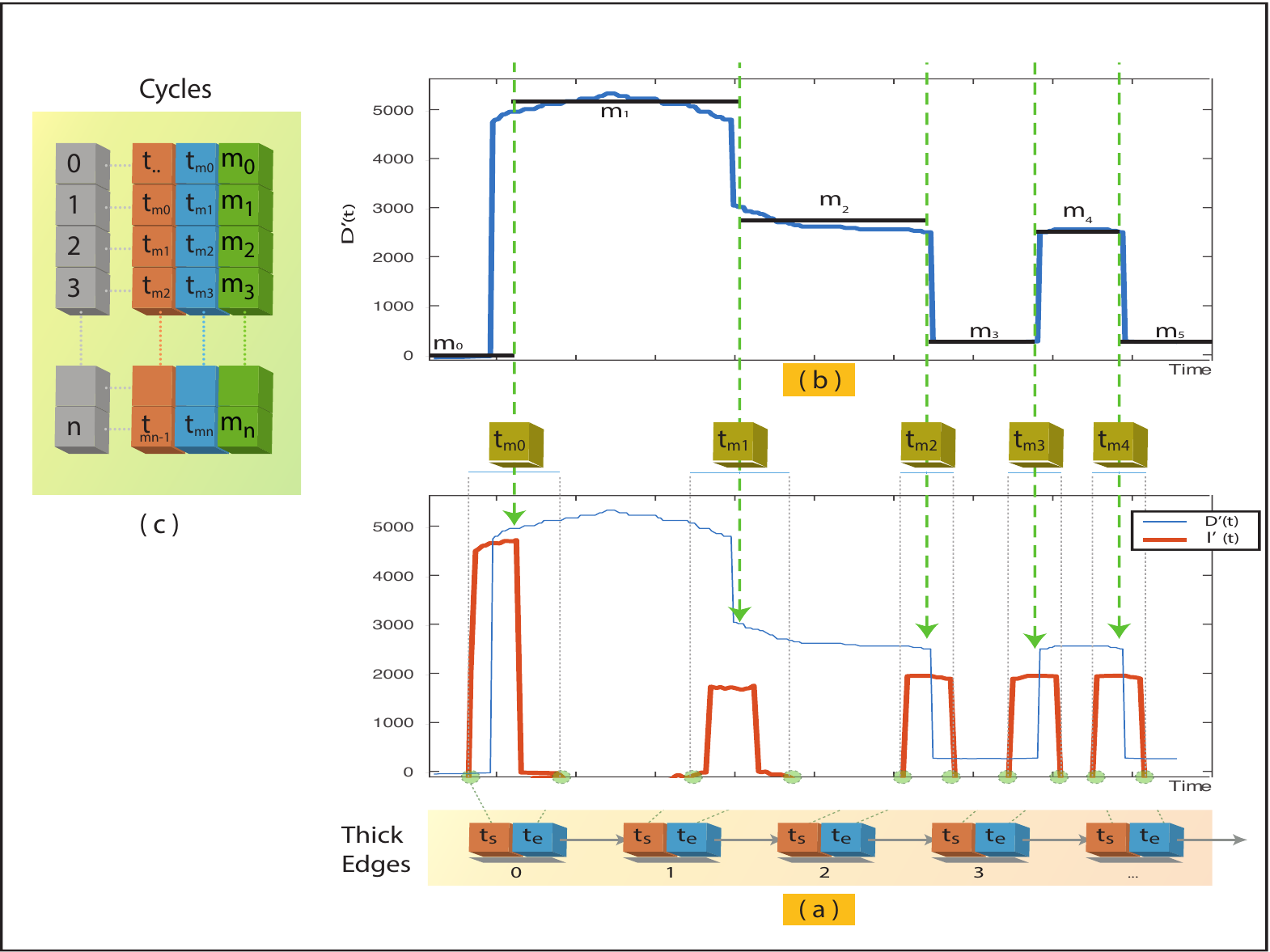}
	\caption{\textbf{(a)} Edge thinning. \textbf{(b)} Extracting the exact edges. \textbf{(c)} The cycles set.}
	\label{fig:mlCycles}
\end{figure}

\cref{fig:mlCycles} (a) demonstrates edge thinning where plots of $I'(t),$ $ D'(t)$ are shown. $E$ is displayed as pairs of times $e_i=(t_{s_i},t_{e_i})$. These time stamps are shown in \cref{fig:mlCycles} (a) as gray dotted lines surrounding each thick edges in both sides. As the definition of the exact edge $t_m$ in \cref{eq:ml05}, exact edges set $X$ are displayed as the yellow time stamps pointing to the middle of each thick edge period with a dashed pointer. 

\paragraph{Extracting Cycles}
To extract the cycles set $C$ of  $D'(t)$, each cycle is defined by two consecutive exact edges $\{t_{m_i}, t_{m_{i+1}}\}$ in addition to the power value of this cycle, $m_i$. To calculate the power value for the $i^{th}$ cycle, the sequence $Y$ is defined as the power values of $D'(t)$ between the selected consecutive exact edges $\{t_{m_i}, t_{m_{i+1}}\}$. We then calculate the cycle power value $m_i$ as the median of $Y$ such that:
\begin{equation} \label{eq:ml07}
m_i  = Median(Y): \quad t\in [t_{m_i}, t_{m_{i+1}}]
\end{equation}   

A demonstration of the cycles extraction is shown in \cref{fig:mlCycles} (b). The exact edges are displayed in yellow boxes pointing upwards towards $D'(t)$ with dashed green lines. The cycle power is defined by the median of $D'(t)$ between two adjacent exact edges. This is depicted as a black horizontal line representing the cycle's power value within the two exact edges, $\{t_{m_i}, t_{m_{i+1}}\}$.

Finally, the cycles set $C$ is formed by collecting what defines a cycle into single tuple, where each tuple $c_i$ hold the start time, end time, and the power value of $c_i$. Therefore, the cycles set $C$ with $n$ number of cycles is defined as the following:

\begin{equation} \label{eq:ml08}
\begin{array}{l}
C = \{c_0, c_1, .. , c_{n-1}\}  \quad\\
c_i = (t_{s_i},t_{e_i},m_i): \quad 0 \leq i < n		
\end{array}
\end{equation} 
where $n$ is the size of $C$ and the $i^{th}$ cycle  is defined by the enclosing two exact edges, $t_{s_i}$ the edge at the cycle start, and $t_{e_i}$ the edge at the cycle end, and the estimated power value of the cycle $m_i$.

\begin{figure}[t]
	\centering
	\includegraphics[width=3.5in]{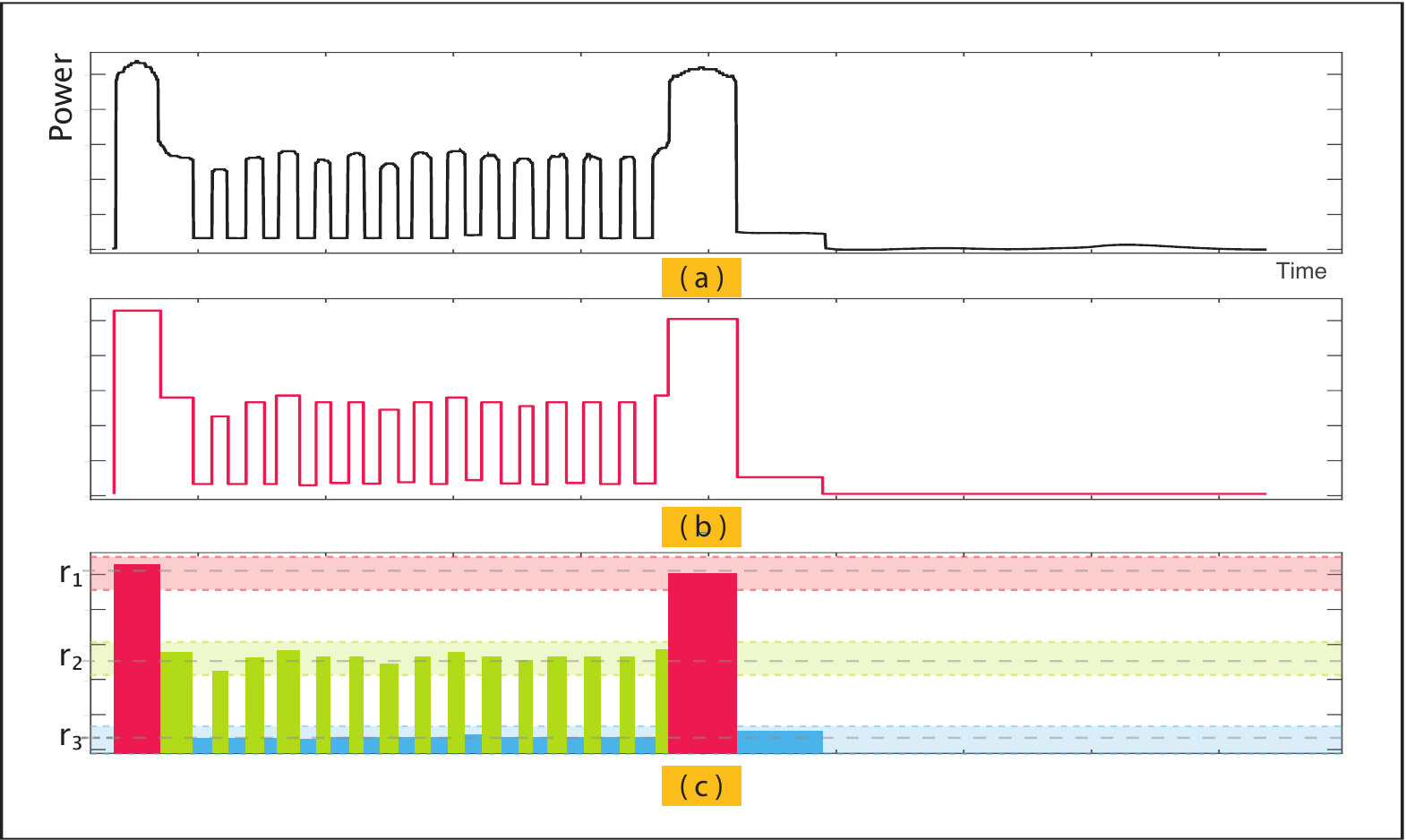}
	\caption{\textbf{(a)} The smoothed function $D'(t)$. \textbf{(b)} A traced version of $D'(t)$ using the cycles extracted. \textbf{(c)} The cycles in different colors based on the power value of each. }
	\label{fig:mlClustering}
\end{figure}

\cref{fig:mlClustering} shows the steps to extract the cycles of a SUP by Edge Detection, Edge Thinning, and Cycle Extraction. \cref{fig:mlClustering} (a) shows the $D'(t)$ after smoothing with the median filter. In \cref{fig:mlClustering} (b) the cycles are extracted into $C$. Using this set, a traced version of $D'(t)$ is sketched based on the information enclosed in the cycles set $C$. Each cycle $c_i$ is traced as a square wave with a magnitude of $m_i$ across the cycle period $[t_{s_i}, t_{e_i}$.

\subsubsection{Cycles Levels Clustering}
As the cycles of a SUP are obtained in $C$, these cycles are used to formulate the features used in the SUP classification. The main objective in this step is to cluster the extracted cycles in $C$ into a set of clusters based on the power level of each cycle.  We selected K-Means Clustering Algorithm \cite[Chapter~20]{mackay2003information} to do this task as K-means is considered a suitable fit for numerical based features \cite{niennattrakul2007clustering}. 

The observations that are fed to K-means consist of the power level $m_i$ for each cycle $c_i \in C$ except the last cycle $e_{n-1}$ as this cycle corresponds to an idle time between SUPs where the power values approaches to zero.  The k-means algorithm requires determining the number of clusters $k$ in advance. Therefore, we select $k=3$ based on the conducted data analysis since the power values of the cycles for all appliances that we analyzed are three power levels. Consequently, K-means produces $k$ mutual exclusive clusters sets $L_0, ... ,  L_{k-1}$ where combining all the clusters represents the observation set such that:

\begin{equation} \label{eq:ml11}
\bigcup_{i=0}^{k-1} L_{i} = L_0 \cup L_1 ... \cup L_{k-1} = B  
\end{equation}

where the observations set $ B=\{m_0, m_1,..., m_{n-1}\}$, and $n$ represents the number of observations.
K-means calculates centroids set $ R=\{r_0, r_1,..., r_i, ..., r_{k-1}\}$, where  $r_i$ is the centroid for the cluster $L_i$ and $k$ is the number of clusters. Each centroids, $r_i$, represents the center point where all the belonging elements $m_j \in L_i$ that have a minimum distance with. 
\cref{fig:mlClustering} (c) shows an example of the cycles in different colors based on the power value of each cycle. Three highlighted areas indicated by the centroid values points $r_0, r_1, r_2 $ show the power range for each cluster.

\begin{figure}[t]
	\centering
	\includegraphics[width=3.5in]{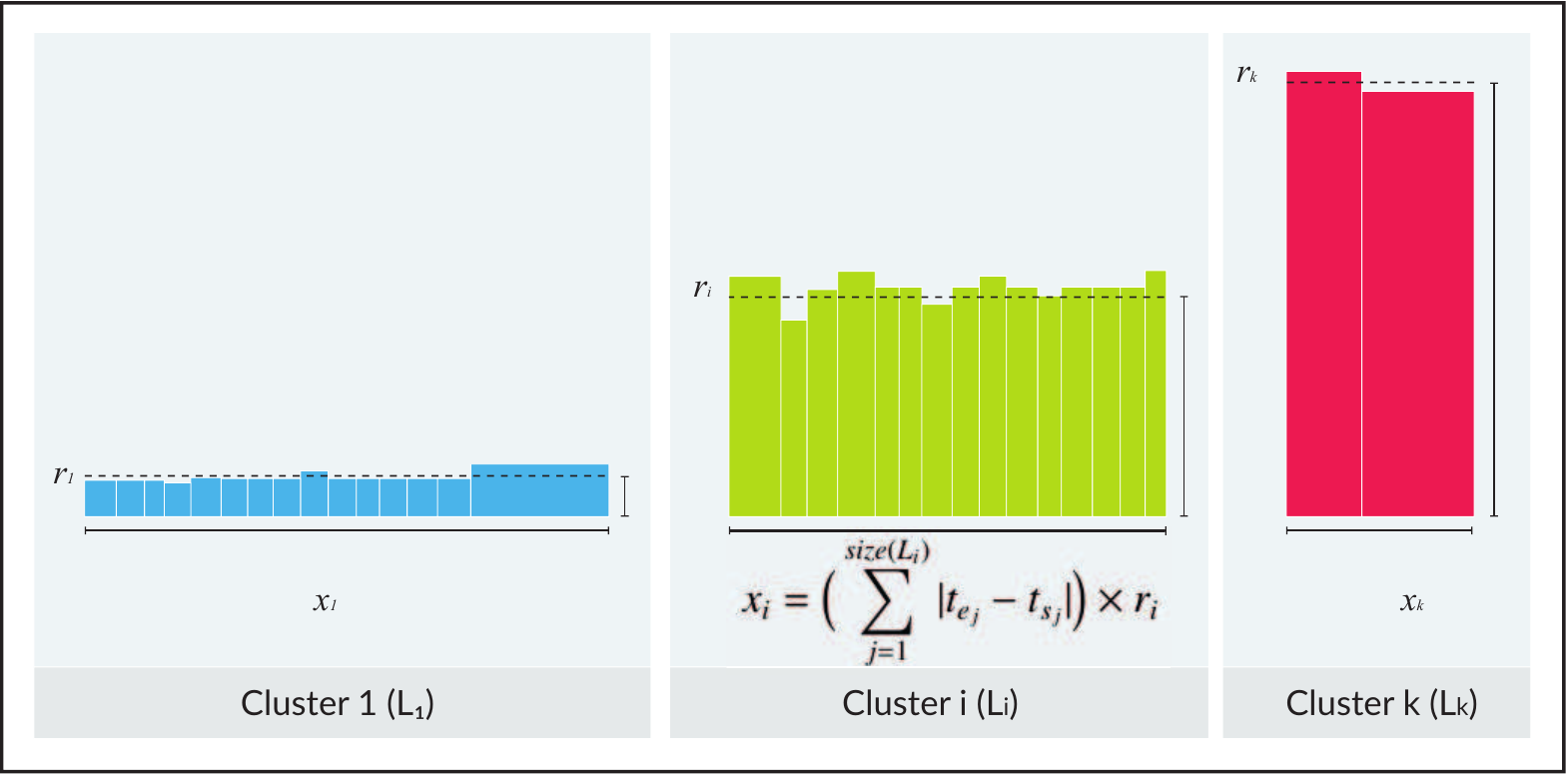}
	\caption{Visual representation for feature calculation.}
	\label{fig:mlFeatures}
\end{figure}

The features set $X$ is calculated from both the clusters in $B$ and the centroids set $R$. Clusters in $B$ are sorted in ascending order based on the centroid value for each cluster such that cluster $L_0$ has the lowest centroid value and cluster $L_{k-1}$ has the highest centroid value. Each cluster $L_i$ is used to define a feature  $x_i$ in the features set $X$. Therefore, the feature set $X$ is defined as:
\begin{equation} \label{eq:ml09}
X = \left\{x_0,x_1, \dots, x_{k-1}\right\}
\end{equation} 
where each feature $x_i$ is defined as the total duration of the cycles $c_j $ within the cluster, $L_i$, multiplied by the average power level of the cluster, which in this case is the value of the centroid of the cluster, $r_i$. Each feature $x_i$ is modeled as follows:
\begin{equation} \label{eq:ml10}
x_i =r_i \sum_{j=0}^{n_i} { | t_{e_j}-t_{s_j} | }    
\end{equation}
where $t_{e_j} , t_{s_j}$ are the two exact edges which define a cycle, as stated in \cref{eq:ml08}, and $n_i$ is the size of $L_i$ . \cref{fig:mlFeatures} visually depicts features formation.

\subsection{Classification Of SUPs To Operation Modes}
\label{sec:KNNClassification}
This section discusses the classification of the SUPs into operation modes based on the features set $X$ that is calculated based on a clustering mechanism for the SUP cycles.

\subsubsection{K-Nearest-Neighbors (KNN) Algorithm}
The algorithm we use in the classification is the K-Nearest-Neighbors (KNN) \cite{biau2015lectures}. It is a classification algorithm that is based on a voting scheme. An item is classified by a plurality voting of its neighbors. The item is then assigned to the class most common between its $ k $ nearest neighbors where $ k>0 $. For example, if $ k = 1 $  the item then is assigned to the class of that single neighbor. 
One of the advantages KNN offers to our study is that KNN is a Lazy Learning Algorithm \cite{punjabi2017lazy}. This means that the generalization of the training set is postponed until a query is made to the algorithm. In other words, KNN keeps track of (at most) all available data all the time until a query of classification is made  so it does the calculations for all the available data with the query. In contrast with the Eager Learning Algorithms, the training set is summarized into a model so that when a query is made the model is enough to do the decision apart form using the training data again. In addition, KNN is suitable for Online Recommendation Systems such as online stores that recommend certain items to the customer \cite{isinkaye2015recommendation}. The reason for its suitability is that the data is continuously updating. As it updates part of the data, other parts my be considered obsolete because of certain trend in the market. Therefore, shrinking the available data into a model and classify upcoming queries based on this model may lead to less accurate results.

\subsubsection{Training Data}
Since we are solving our main problem using a classification technique, it is necessary to have a dataset that is used to train the classifier. Due to the nature of the KNN algorithm as a lazy learning algorithm, it is important to initialize the KNN model and compute initial values for its centroids so that it can work online. This initialization is considered as a kind of training stage for the KNN model which can not be obtained without an pre-labeled dataset.

The architecture demonstrated in \cref{fig:mlModel} shows that the classification module uses synthetic data based on existing sets. The simulator generates a synthetic dataset that consists of all possibilities of operation modes for appliance SUP with variations through changing tuning parameters that makes the dataset diverse. The simulator is already configured with distribution functions of the SUP cycles in terms of duration, power level, and repetition. The simulator then synthesizes a set of feature vectors that match the definition of \cref{eq:ml10}, and is made available to the classifier.

\begin{figure}[t]
	\centering
	\fbox{\includegraphics[trim={1.7cm 1.7cm 1.7cm 1.7cm },clip,width=.475\textwidth]{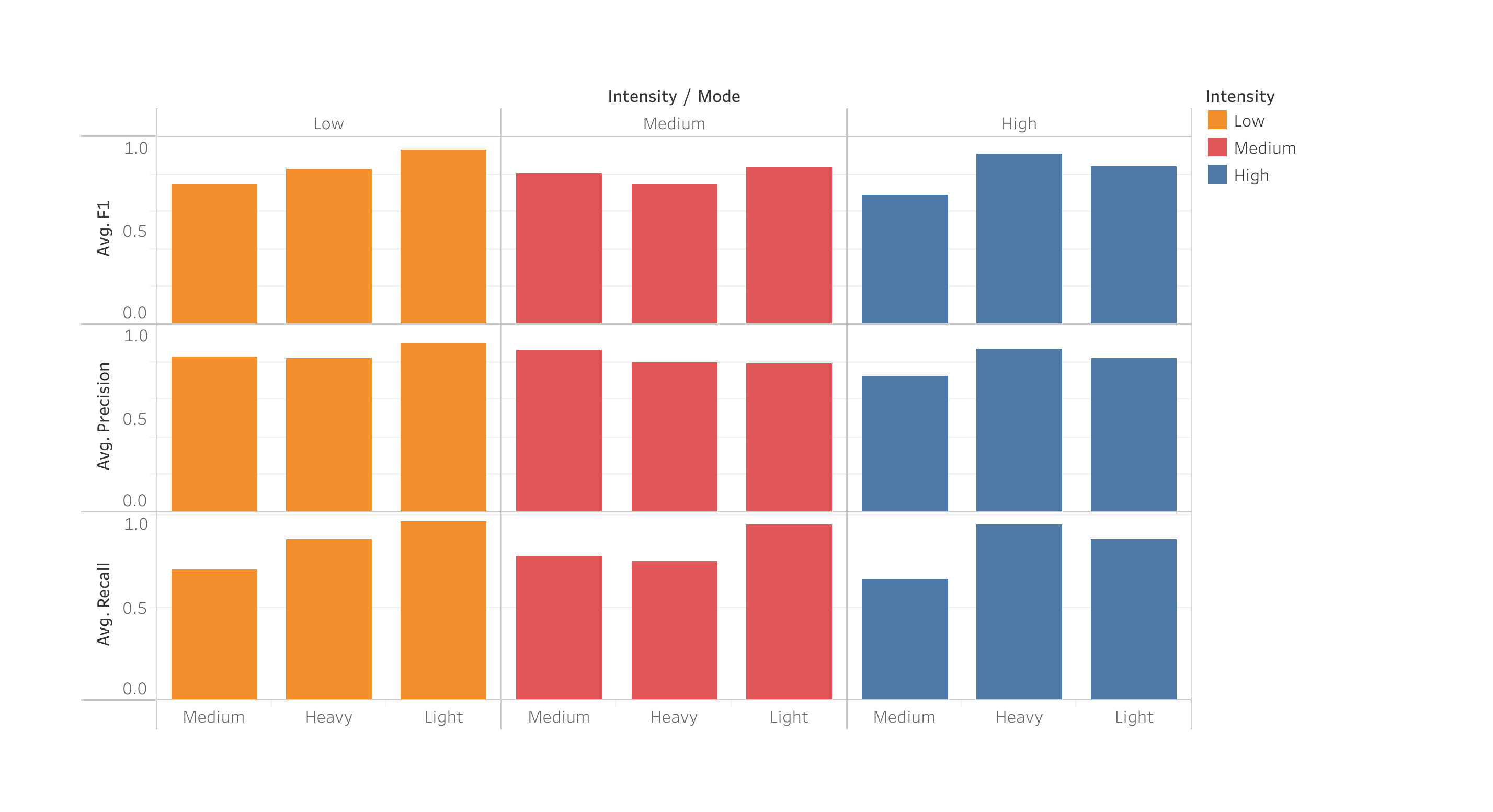}}
	\caption{The precision, recall, and F1-score for each operation mode in every usage intensity.}
	\label{fig:evalIntensity}
\end{figure}

The training data is a synthetic dataset generated by Power Consumption Simulator (PCS) \cite{jaradat2020demand} . This dataset consists of three subsets, where each subset consists of 4000 observations for certain appliance (dishwasher, clothes dryer, clothes washer). The training set consists of a set of observations where each observation represents a single SUP as a set of features $X$ with size $k$ as defined in \cref{eq:ml09}.
As the feature set is extracted, the features set with $k$ features takes the form $	X = \{x_0,x_1, .., x_i, .., x_{k-1}\}$. The training and features sets are fed to the KNN classifier in order to classify each SUP into one of $n$ operation modes, $d_i$, within the set of operation modes $M=\{d_0, d_1,...,d_{n-1}\}$. In our study, the focus was on $n=3$ operation modes as $M=\{Light, Medium, Heavy\}$.
\section{Results And Discussion}

 \begin{figure}[t]
	\centering

	\includegraphics[width=.475\textwidth]{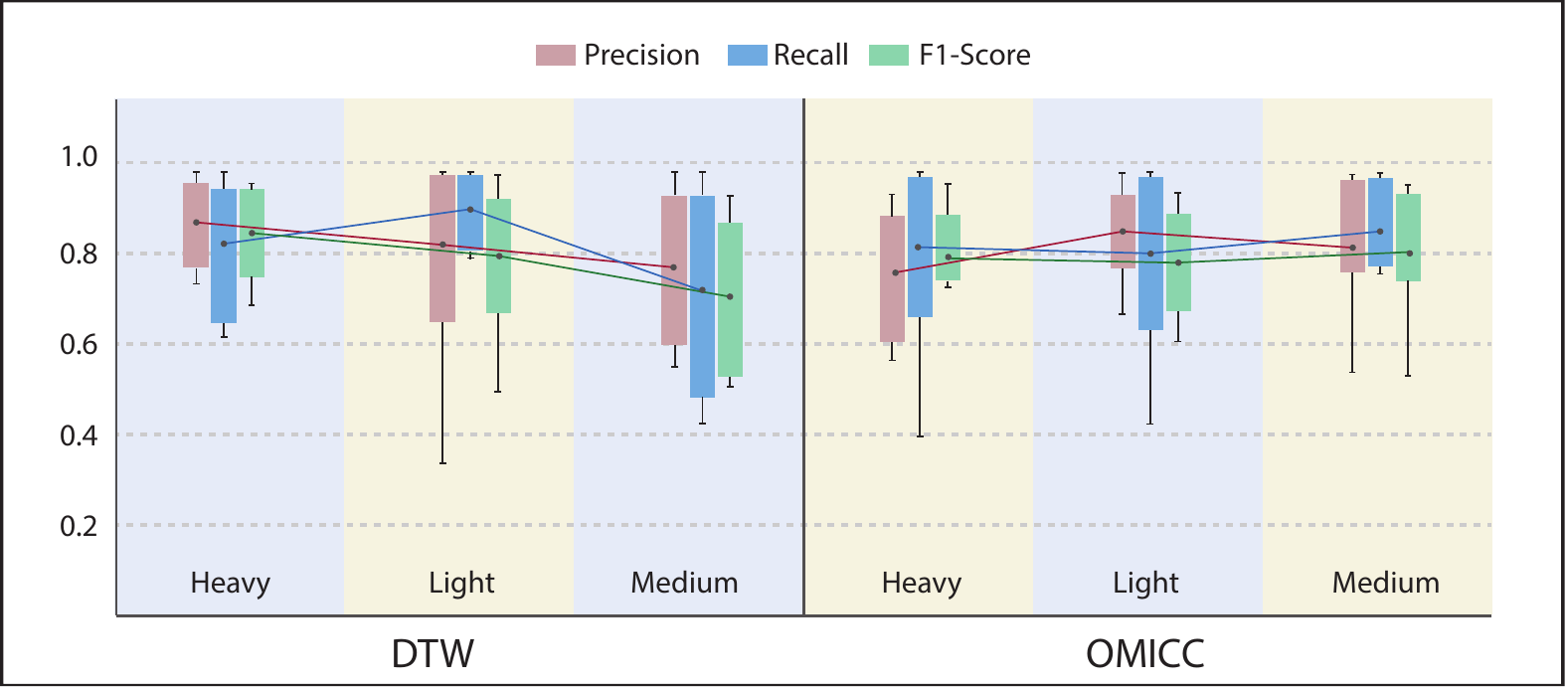}
	\caption{The precision, recall and F1-score for each operation mode using OMICC compared to  DTW. }
	\label{fig:DTWbox}
	
\end{figure}

The performance for OMICC is evaluated using supervised learning classification metrics i.e., Precision, Recall, and F1-score. The classification of SUPs in OMICC is compared to our previous work by classifying SUPs using Dynamic Time Warping (DTW) classification \cite{jaradat2020demand}.
The performance of the DTW approach and OMICC  is summarized in \cref{table:evalAlgPrecRec}. The table shows the average precision, recall, and the F1-score across all the experiments conducted over the datasets. The precision and recall values of the DTW are 81\% and 81\% respectively. While these values are generally higher for OMICC with 84\% and 85\%.

\begin{table}[ht]
	\centering
		\caption{Average precision, recall, and F1-score values for all datasets using DTW based \cite{jaradat2020demand} and OMICC.}
	\begin{tabularx}{.475\textwidth}{XXXX}
		\bottomrule\hline
		\textbf{Algorithm}	&\textbf{Precision} & \textbf{Recall}	& \textbf{F1-score} \\\hline
		\textbf{DTW} &	0.81 &	0.81 &	0.80 	\\
		\textbf{OMICC} &	0.84 &	0.85 &	0.83 	\\	
		\bottomrule\hline
	\end{tabularx}

	\label{table:evalAlgPrecRec}
\end{table}

\cref{fig:evalIntensity} shows the relationship between the Household Usage Intensity Distribution, the operation mode, and the metrics. Usage intensity refers to the pattern of operation modes that the household selects over time for appliances \cite{jaradat2020demand}. The chart shows the precision, recall, and F1-score on the y-axis. The x-axis is divided into three lanes representing the usage intensity values (low, medium, high). In each lane, the operation mode is depicted. The chart shows high values of the metric for the major operation mode in each usage intensity lane. e.g., in the low intensity lane, the light mode is the operation mode that is used by the household the majority of the time. Therefore, the three metric values for the light mode are relatively higher than other modes. The heavy operation mode in the high intensity lane behaves in the same way. The medium operation mode in the medium intensity lane show this property in the precision, but generally it has higher performance than other usage intensity.

The evaluation of the performance of the two algorithms used in terms of operation modes is depicted in \cref{fig:DTWbox}. The charts shows the precision, recall and F1-score for each operation mode using box and whiskers. \cref{fig:DTWbox} shows that for the DTW the metrics values are averaged around 82\% for the light and heavy operation modes. Otherwise, the metrics value is approximately 79\%. On the other hand, the KNN chart presented in \cref{fig:DTWbox} shows more balanced results, for the three operation modes the metrics average values are around 82\%.

\section{Conclusion And Future Work}
\label{conc}
Our work is focused on providing techniques built on top of residential power consumption to better support DR. We proposed Operation Modes Identification using Cycles Clustering (OMICC) which is SHEMS fundamental approach that processes residential disaggregated power consumption data to support DR. 
The cycles for SUPs from appliances are extracted using MST and clustered using K-means to form features that represent the SUP. We then utilized KNN to classify SUPs and identify the corresponding operation modes accordingly. The identified operation modes gives insights to the consumer on mode selection, and hence energy consumption reduction through choosing lighter operation modes.
A future improvement to the current work is to enhance the SUP classification by taking into consideration the cycles order within SUP in feature extraction. Furthermore, proposing a SHEMS on top of OMICC to expose its applicability by reporting the identified operation modes to consumers and offering opportunities for energy conservation.

\ifCLASSOPTIONcaptionsoff
  \newpage
\fi



%
\bibliographystyle{plain} 

\bibliography{tex/bib/library,tex/bib/others,tex/bib/web,tex/bib/bib}{}

%





\end{document}